# Query DAGs: A Practical Paradigm for Implementing Belief-Network Inference

**Adnan Darwiche**                                    DARWICHE@AUB.EDU.LB
*Department of Mathematics*
*American University of Beirut*
*PO Box 11 - 236, Beirut, Lebanon*

**Gregory Provan**                                    PROVAN@RISC.ROCKWELL.COM
*Rockwell Science Center*
*1049 Camino Dos Rios*
*Thousand Oaks, CA 91360*

## Abstract

We describe a new paradigm for implementing inference in belief networks, which consists of two steps: (1) compiling a belief network into an arithmetic expression called a Query DAG (Q-DAG); and (2) answering queries using a simple evaluation algorithm. Each node of a Q-DAG represents a numeric operation, a number, or a symbol for evidence. Each leaf node of a Q-DAG represents the answer to a network query, that is, the probability of some event of interest. It appears that Q-DAGs can be generated using any of the standard algorithms for exact inference in belief networks — we show how they can be generated using clustering and conditioning algorithms. The time and space complexity of a Q-DAG *generation* algorithm is no worse than the time complexity of the inference algorithm on which it is based. The complexity of a Q-DAG *evaluation* algorithm is linear in the size of the Q-DAG, and such inference amounts to a standard evaluation of the arithmetic expression it represents. The intended value of Q-DAGs is in reducing the software and hardware resources required to utilize belief networks in on-line, real-world applications. The proposed framework also facilitates the development of on-line inference on different software and hardware platforms due to the simplicity of the Q-DAG evaluation algorithm. Interestingly enough, Q-DAGs were found to serve other purposes: simple techniques for reducing Q-DAGs tend to subsume relatively complex optimization techniques for belief-network inference, such as network-pruning and computation-caching.

## 1. Introduction

Consider designing a car to have a self-diagnostic system that can alert the driver to a range of problems. Figure 1 shows a simplistic belief network that could provide a ranked set of diagnoses for car troubleshooting, given input from sensors hooked up to the battery, alternator, fuel-tank and oil-system.

The standard approach to building such a diagnostic system is to put this belief network, along with inference code, onto the car's computer; see Figure 2. We have encountered a number of difficulties when using this approach to embody belief network technology in industrial applications. First, we were asked to provide the technology on multiple platforms. For some applications, the technology had to be implemented in ADA to pass certain certification procedures. In others, it had to be implemented on domain-specific hardware that only supports very primitive programming languages. Second, memory was limited to keep the cost of a unit below a certain threshold to maintain product profitability. The dilemma was the following: belief network algorithms are not trivial to implement, especially when optimization is crucial, and porting these algorithms to multiple platforms and languages would have been prohibitively expensive, time-consuming and demanding of qualified manpower.

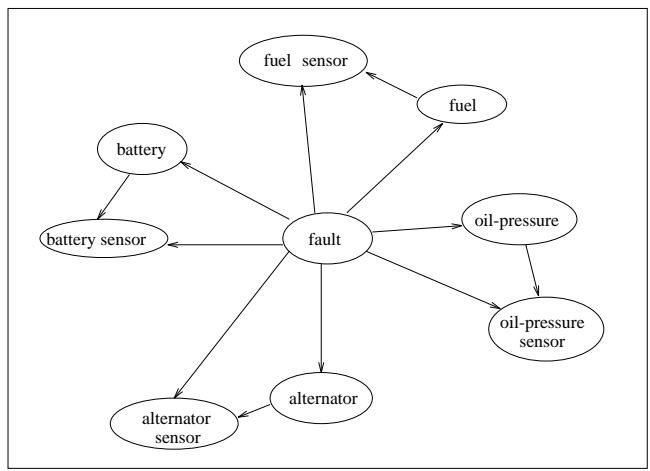

Figure 1: A simple belief network for car diagnosis.

To overcome these difficulties, we have devised a very flexible approach for implementing belief network systems, which is based on the following observation. Almost all the work performed by standard algorithms for belief networks is independent of the specific evidence gathered about variables. For example, if we run an algorithm with the battery-sensor set to *low* and then run it later with the variable set to *dead*, we find almost no algorithmic difference between the two runs. That is, the algorithm will not branch differently on any of the key decisions it makes, and the only difference between the two runs is the specific arguments to the invoked numeric operations. Therefore, one can apply a standard inference algorithm on a network with evidence being a *parameter* instead of being a specific value. The result returned by the algorithm will then be an arithmetic expression with some parameters that depend on specific evidence. This parameterized expression is what we call a Query DAG, an example of which is shown in Figure 4.[1]

The approach we are proposing consists of two steps. First, given a belief network, a set of variables about which evidence may be collected (evidence variables), and a set of variables for which we need to compute probability distributions (query variables), a Q-DAG is compiled off-line, as shown in Figure 3. The compilation is typically done on a sophisticated software/hardware platform, using a traditional belief network inference algorithm in conjunction with the Q-DAG compilation method. This part of the process is far and away the most costly computationally. Second, an on-line system composed from the generated Q-DAG and an evaluator specific to the given platform is used to evaluate the Q-DAG. Given evidence, the parameterized arithmetic expression is evaluated in a straightforward manner using simple arithmetic operations rather than complicated belief network inference. The computational work needed to perform this on-line evaluation is so straightforward that it lends itself to easy implementations on different software and hardware platforms.

---

1. The sharing of subexpressions is what makes this a Directed Acyclic Graph instead of a tree.

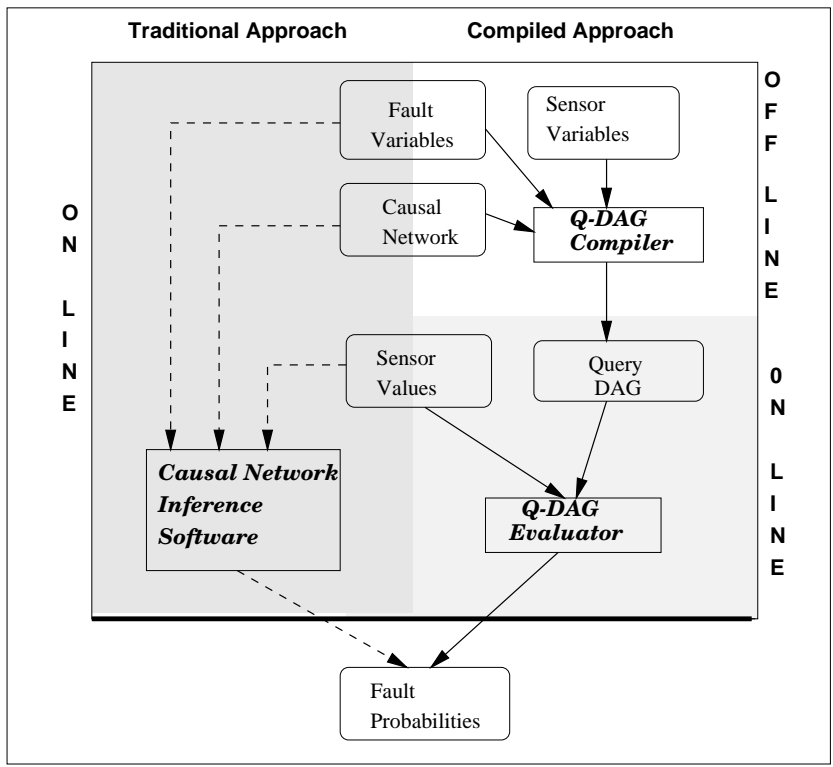

Figure 2: This figure compares the traditional approach to exact belief-network inference (shown on the left) with our new compiled approach (shown on the right) in the context of diagnostic reasoning. In the traditional approach, the belief network and sensor values are used *on-line* to compute the probability distributions over fault variables; in the compiled approach, the belief network, fault variables and sensor variables are compiled *off-line* to produce a Q-DAG, which is then evaluated *on-line* using sensor values to compute the required distributions.

This approach shares some commonality with other methods that symbolically manipulate probability expressions, like SPI (Li & D'Ambrosio, 1994; Shachter, D'Ambrosio, & del Favero, 1990); it differs from SPI on the objective of such manipulations and, hence, on the results obtained. SPI explicates the notion of an arithmetic expression to state that belief-network inference can be viewed as an expression-factoring operation. This allows results from optimization theory to be utilized in belief-network inference. On the other hand, we define an arithmetic expression to explicate and formalize the boundaries between on-line and off-line inference, with the goal of identifying the minimal piece of software that is required on-line. Our results are therefore oriented towards this purpose and they include: (a) a formal definition of a Q-DAG and its evaluator; (b) a method for generating Q-DAGs using standard inference algorithms — an algorithm need not subscribe to the inference-as-factoring view to be used for Q-DAG generation; and (c) computational guarantees on the size of Q-DAGs in terms of the computational guarantees of the inference algorithm used to generate them. Although the SPI framework is positioned to formulate related results, it has not been pursued in this direction.

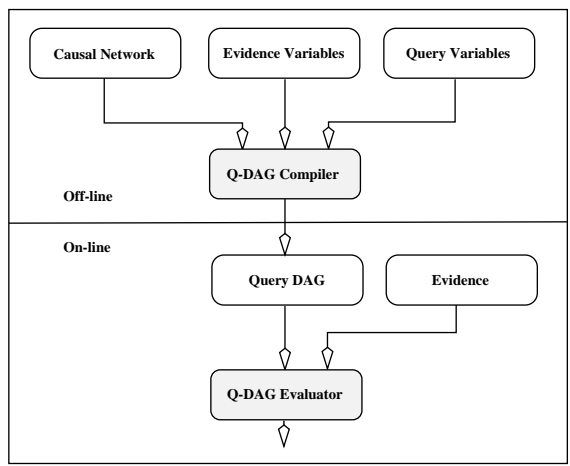

Figure 3: The proposed framework for implementing belief-network inference.

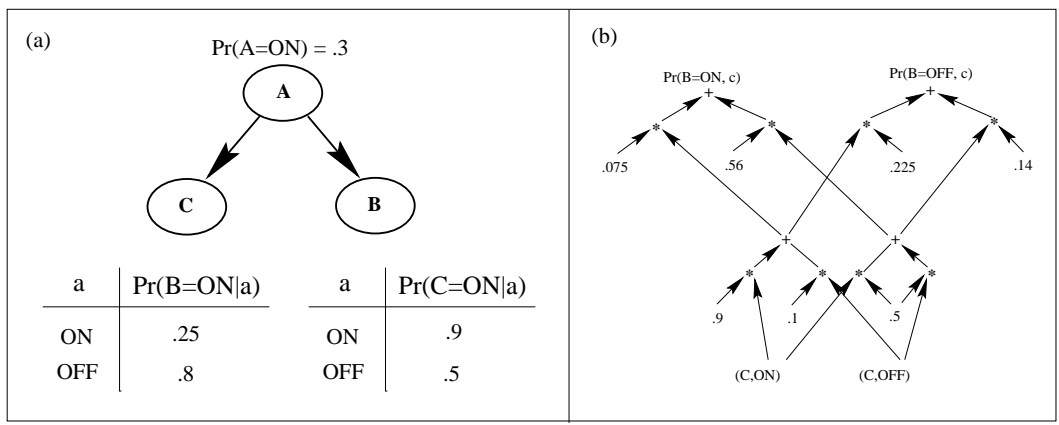

Figure 4: A belief network (a); and its corresponding Query-DAG (b). Here, $C$ is an evidence variable, and we are interested in the probability of variable $B$.

It is important to stress the following properties of the proposed approach. First, declaring an evidence variable in the compilation process does *not* mean that evidence must be collected about that variable on-line—this is important because some evidence values, e.g., from sensors, may be lost in practice—it only means that evidence *may* be collected. Therefore, one can declare all variables to be evidence if one wishes. Second, a variable can be declared to be both evidence and query. This allows one to perform value-of-information computations to decide whether it is worth collecting evidence about a specific variable. Third, the space complexity of a Q-DAG in terms of the number of evidence variables is no worse than the time complexity of its underlying inference algorithm; therefore, this is not a simple enumerate-all-possible-cases approach. Finally, the time and space complexity for generating a Q-DAG is no worse than the time complexity of the standard belief-network algorithm used in its generation. Therefore, if a network can be solved using a standard inference algorithm, and if the time complexity of this algorithm is no worse than its space complexity,[2] then we can construct a Q-DAG for that network.

The following section explains the concept of a Q-DAG with a concrete example and provides formal definitions. Section 3 is dedicated to the generation of Q-DAGs and their computational complexity, showing that any standard belief-network inference algorithm can be used to compile a Q-DAG as long as it meets some general conditions. Section 4 discusses the reduction of a Q-DAG after it has been generated, showing that such reduction subsumes key optimizations that are typically implemented in belief network algorithms. Section 5 contains a detailed example on the application of this framework to diagnostic reasoning. Finally, Section 6 closes with some concluding remarks.

## 2. Query DAGs

This section starts our treatment of Q-DAGs with a concrete example. We will consider a particular belief network, define a set of queries of interest, and then show a Q-DAG that can be used to answer such queries. We will not discuss how the Q-DAG is generated; only how it can be used. This will allow a concrete introduction to Q-DAGs and will help us ground some of the formal definitions to follow.

The belief network we will consider is the one in Figure 4(a). The class of queries we are interested in is $Pr(B \mid C)$, that is, the probability that variable $B$ takes some value given some known (or unknown) value of $C$. Figure 4(b) depicts a Q-DAG for answering such queries, which is essentially a parameterized arithmetic expression where the values of parameters depend on the evidence obtained. This Q-DAG will actually answer queries of the form $Pr(B, C)$, but we can use normalization to compute $Pr(B \mid C)$.

First, a number of observations about the Q-DAG in Figure 4(b):

- The Q-DAG has two leaf nodes labeled $Pr(B{=}ON, c)$ and $Pr(B{=}OFF, c)$. These are called *query nodes* because their values represent answers to the queries $Pr(B{=}ON, c)$ and $Pr(B{=}OFF, c)$.

- The Q-DAG has two root nodes labeled $(C, ON)$ and $(C, OFF)$. These are called *Evidence Specific Nodes (ESNs)* since their values depend on the evidence collected about variable $C$ on-line.

According to the semantics of Q-DAGs, the value of node $(V, v)$ is 1 if variable $V$ is observed to be $v$ or is unknown, and 0 otherwise. Once the values of ESNs are determined, we evaluate the remaining nodes of a Q-DAG using numeric multiplication and addition. The numbers that get assigned to query nodes as a result of this evaluation are the answers to queries represented by these nodes.

---

2. Algorithms based on join trees have this property.

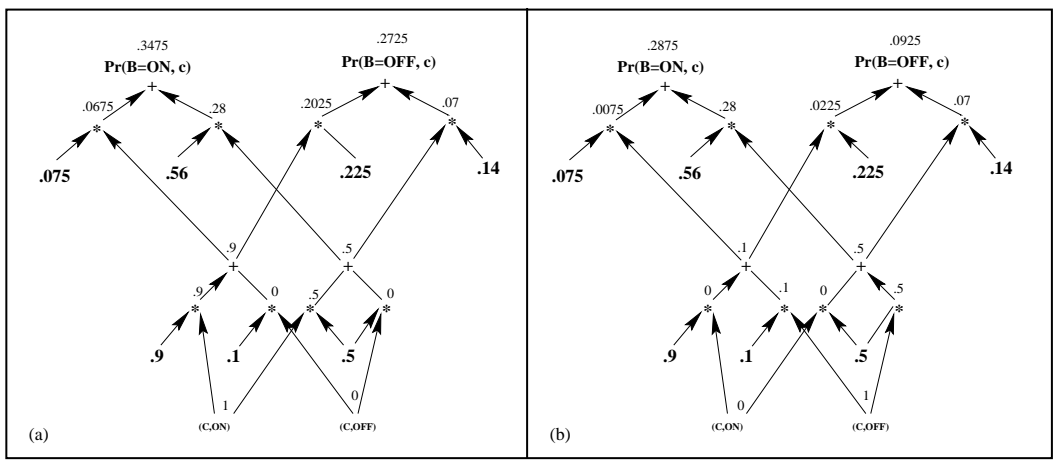

Figure 5: Evaluating the Q-DAG in Figure 4 with respect to two pieces of evidence: (a) $C=ON$ and (b) $C=OFF$.

For example, suppose that the evidence we have is $C = ON$. Then ESN $(C, ON)$ is evaluated to 1 and ESN $(C, OFF)$ is evaluated to 0. The Q-DAG in Figure 4(b) is then evaluated as given in Figure 5(a), thus leading to

$$Pr(B=ON, C=ON) = .3475,$$

and

$$Pr(B=OFF, C=ON) = .2725,$$

from which we conclude that $Pr(C=ON) = .62$. We can then compute the conditional probabilities $Pr(B=ON \mid C=ON)$ and $Pr(B=OFF \mid C=ON)$ using:

$$Pr(B=ON \mid C=ON) = Pr(B=ON, C=ON)/Pr(C=ON),$$

$$Pr(B=OFF \mid C=ON) = Pr(B=OFF, C=ON)/Pr(C=ON).$$

If the evidence we have is $C=OFF$, however, then $(C, ON)$ evaluates to 0 and $(C, OFF)$ evaluates to 1. The Q-DAG in Figure 4(b) will then be evaluated as given in Figure 5(b), thus leading to

$$Pr(B=ON, C=OFF) = .2875,$$

and

$$Pr(B=OFF, C=OFF) = .0925.$$

We will use the following notation for denoting variables and their values. Variables are denoted using uppercase letters, such as $A, B, C$, and variable values are denoted by lowercase letters, such as $a, b, c$. Sets of variables are denoted by boldface uppercase letters, such as $\mathbf{A}, \mathbf{B}, \mathbf{C}$, and their instantiations are denoted by boldface lowercase letters, such as $\mathbf{a}, \mathbf{b}, \mathbf{c}$. We use $\mathbf{E}$ to denote the set of variables about which we have evidence. Therefore, we use **e** to denote an instantiation of these variables that represents evidence. Finally, the *family* of a variable is the set containing the variable and its parents in a directed acyclic graph.

Following is the formal definition of a Q-DAG.

**Definition 1** *A Q-DAG is a tuple* $(\mathcal{V}, \diamond, \mathcal{I}, \mathcal{D}, \mathcal{Z})$ *where*

1. $\mathcal{V}$ *is a distinguished set of symbols (called <u>evidence variables</u>)*

2. $\diamond$ *is a symbol (called <u>unknown value</u>)*

3. $\mathcal{I}$ *maps each variable in* $\mathcal{V}$ *into a set of symbols (called <u>variable values</u>) different from* $\diamond$.

4. $\mathcal{D}$ *is a directed acyclic graph where*

   - *each non-root node is labeled with either* $+$ *or* $*$
   - *each root node is labeled with either*

     - *a number in* $[0,1]$ *or*
     - *a pair* $(V, v)$ *where* $V$ *is an evidence variable and* $v$ *is a value*

5. $\mathcal{Z}$ *is a distinguished set of nodes in* $\mathcal{D}$ *(called <u>query nodes</u>)*

Evidence variables $\mathcal{V}$ correspond to network variables about which we expect to collect evidence on-line. For example, in Figure 5, $C$ is the evidence variable. Each one of these variables has a set of possible values that are captured by the function $\mathcal{I}$. For example, in Figure 5, the evidence variable $C$ has values $ON$ and $OFF$. The special value $\diamond$ is used when the value of a variable is not known. For example, we may have a sensor variable with values "low," "medium," and "high," but then lose the sensor value during on-line reasoning. In this case, we set the sensor value to $\diamond$.[3] Query nodes are those representing answers to user queries. For example, in Figure 5, $B$ is the query variable, and leads to query nodes $Pr(B{=}ON, c)$ and $Pr(B{=}OFF, c)$.

An important notion is that of evidence:

**Definition 2** *For a given Q-DAG* $(\mathcal{V}, \diamond, \mathcal{I}, \mathcal{D}, \mathcal{Z})$, <u>evidence</u> *is defined as a function* $\mathcal{E}$ *that maps each variable* $V$ *in* $\mathcal{V}$ *into the set of values* $\mathcal{I}(V) \cup \{\diamond\}$.

When a variable $V$ is mapped into $v \in \mathcal{I}(V)$, then evidence tells us that $V$ is instantiated to value $v$. When $V$ is mapped into $\diamond$, then evidence does not tell us anything about the value of $V$.

We can now state formally how to evaluate a Q-DAG given some evidence. But first we need some more notation:

1. *Numeric-Node:* $n(p)$ denotes a node labeled with a number $p \in [0,1]$;

2. *ESN:* $n(V, v)$ denotes a node labeled with $(V, v)$;

---

3. This is also useful in cases where a variable will be measured only if its value of information justifies that.

3. *Operation-Node:* $n_1 \otimes \ldots \otimes n_i$ denotes a node labeled with $*$ and having parents $n_1, \ldots, n_i$;

4. *Operation-Node:* $n_1 \oplus \ldots \oplus n_i$ denotes a node labeled with $+$ and having parents $n_1, \ldots, n_i$.

The following definition tells us how to evaluate a Q-DAG by evaluating each of its nodes. It is a recursive definition according to which the value assigned to a node is a function of the values assigned to its parents. The first two cases are boundary conditions, assigning values to root nodes. The last two cases are the recursive ones.

**Definition 3** *For a Q-DAG $(\mathcal{V}, \diamond, \mathcal{I}, \mathcal{D}, \mathcal{Z})$ and evidence $\mathcal{E}$, the <u>node evaluator</u> is defined as a function $\mathcal{M}_\mathcal{E}$ that maps each node in $\mathcal{D}$ into a number $[0, 1]$ such that:*

1. $\mathcal{M}_\mathcal{E}[n(p)] = p$
   *(The value of a node labeled with a number is the number itself.)*

2. $\mathcal{M}_\mathcal{E}[n(V, v)] = \begin{cases} 1, & \text{if } \mathcal{E}(V) = v \text{ or } \mathcal{E}(V) = \diamond; \\ 0, & \text{otherwise} \end{cases}$
   *(The value of an evidence-specific node depends on the available evidence: it is 1 if $v$ is consistent with the evidence and 0 otherwise.)*

3. $\mathcal{M}_\mathcal{E}[n_1 \otimes \ldots \otimes n_i] = \mathcal{M}_\mathcal{E}(n_1) * \ldots * \mathcal{M}_\mathcal{E}(n_i)$
   *(The value of a node labeled with $*$ is the product of the values of its parent nodes.)*

4. $\mathcal{M}_\mathcal{E}[n_1 \oplus \ldots \oplus n_i] = \mathcal{M}_\mathcal{E}(n_1) + \ldots + \mathcal{M}_\mathcal{E}(n_i)$
   *(The value of a node labeled with $+$ is the sum of the values of its parent nodes.)*

One is typically not interested in the values of all nodes in a Q-DAG since most of these nodes represent intermediate results that are of no interest to the user. It is the query nodes of a Q-DAG that represent answers to user queries and it is the values of these nodes that one seeks when constructing a Q-DAG. The values of these queries are captured by the notion of a Q-DAG output.

**Definition 4** *The node evaluator $\mathcal{M}_\mathcal{E}$ is extended to Q-DAGs as follows:*

$$\mathcal{M}_\mathcal{E}((\mathcal{V}, \diamond, \mathcal{I}, \mathcal{D}, \mathcal{Z})) = \{(n, \mathcal{M}_\mathcal{E}(n)) \mid n \in \mathcal{Z}\}.$$

*The set $\mathcal{M}_\mathcal{E}((\mathcal{V}, \diamond, \mathcal{I}, \mathcal{D}, \mathcal{Z}))$ is called the Q-DAG <u>output</u>.*

This output is what one seeks from a Q-DAG. Each element in this output represents a probabilistic query and its answer.

Let us consider a few evaluations of the Q-DAG shown in Figure 4, which are shown in Figure 5. Given evidence $\mathcal{E}(C) = ON$, and assuming that $Qnode(B = ON)$ and $Qnode(B = OFF)$ stand for the Q-DAG nodes labeled $Pr(B = ON, c)$ and $Pr(B = OFF, c)$, respectively, we have

$$
\begin{aligned}
\mathcal{M}_\mathcal{E}[n(C, ON)] &= 1, \\
\mathcal{M}_\mathcal{E}[n(C, OFF)] &= 0, \\
\mathcal{M}_\mathcal{E}[Qnode(B = ON)] &= .075 * (.9 * 1 + .1 * 0) + .56 * (1 * .5 + .5 * 0) = .3475, \\
\mathcal{M}_\mathcal{E}[Qnode(B = OFF)] &= (.9 * 1 + .1 * 0) * .225 + (1 * .5 + .5 * 0) * .14 = .2725,
\end{aligned}
$$

meaning that $Pr(B{=}ON, C{=}ON) = .3475$ and $Pr(B{=}OFF, C{=}ON) = .2725$. If instead the evidence were $\mathcal{E}(C){=}OFF$, a set of analogous computations can be done.

It is also possible that evidence tells us nothing about the value of variable $C$, that is, $\mathcal{E}(C) = \diamond$. In this case, we would have

$$
\begin{aligned}
\mathcal{M}_{\mathcal{E}}[n(C, ON)] &= 1, \\
\mathcal{M}_{\mathcal{E}}[n(C, OFF)] &= 1, \\
\mathcal{M}_{\mathcal{E}}[Qnode(B{=}ON)] &= .075*(.9*1+.1*1)+.56*(1*.5+.5*1) = .635, \\
\mathcal{M}_{\mathcal{E}}[Qnode(B{=}OFF)] &= (.9*1+.1*1)*.225+(1*.5+.5*1)*.14 = .365,
\end{aligned}
$$

meaning that $Pr(B{=}ON) = .635$ and $Pr(B{=}OFF) = .365$.

## 2.1 Implementing a Q-DAG Evaluator

A Q-DAG evaluator can be implemented using an event-driven, forward propagation scheme. Whenever the value of a Q-DAG node changes, one updates the value of its children, and so on, until no possible update of values is possible. Another way to implement an evaluator is using a backward propagation scheme where one starts from a query node and updates its value by updating the values of its parent nodes. The specifics of the application will typically determine which method (or combination) will be more appropriate.

It is important that we stress the level of refinement enjoyed by the Q-DAG propagation scheme and the implications of this on the efficiency of query updates. Propagation in Q-DAGs is done at the arithmetic-operation level, which is contrasted with propagation at the message-operation level (used by many standard algorithms). Such propagation schemes are typically optimized by keeping validity flags of messages so that only invalid messages are recomputed when new evidence arrives. This will clearly avoid some unnecessary computations but can never avoid all unnecessary computations because a message is typically too coarse for this purpose. For example, if only one entry in a message is invalid, the whole message is considered invalid. Recomputing such a message will lead to many unnecessary computations. This problem will be avoided in Q-DAG propagation since validity flags are attributed to *arithmetic* operations, which are the building blocks of message operations. Therefore, only the necessary arithmetic operations will be recomputed in a Q-DAG propagation scheme, leading to a more detailed level of optimization.

We also stress that the process of evaluating and updating a Q-DAG is done outside of probability theory and belief network inference. This makes the development of efficient online inference software accessible to a larger group of people who may lack strong backgrounds in these areas.[4]

## 2.2 The Availability of Evidence

The construction of a Q-DAG requires the identification of query and evidence variables. This may give an incorrect impression that we must know up front which variables are observed and which are not. This could be problematic in (1) applications where one may lose a sensor reading, thus changing the status of a variable from being observed to being unobserved;

---

4. In fact, it appears that a background in compiler theory may be more relevant to generating an efficient evaluator than a background in belief network theory.

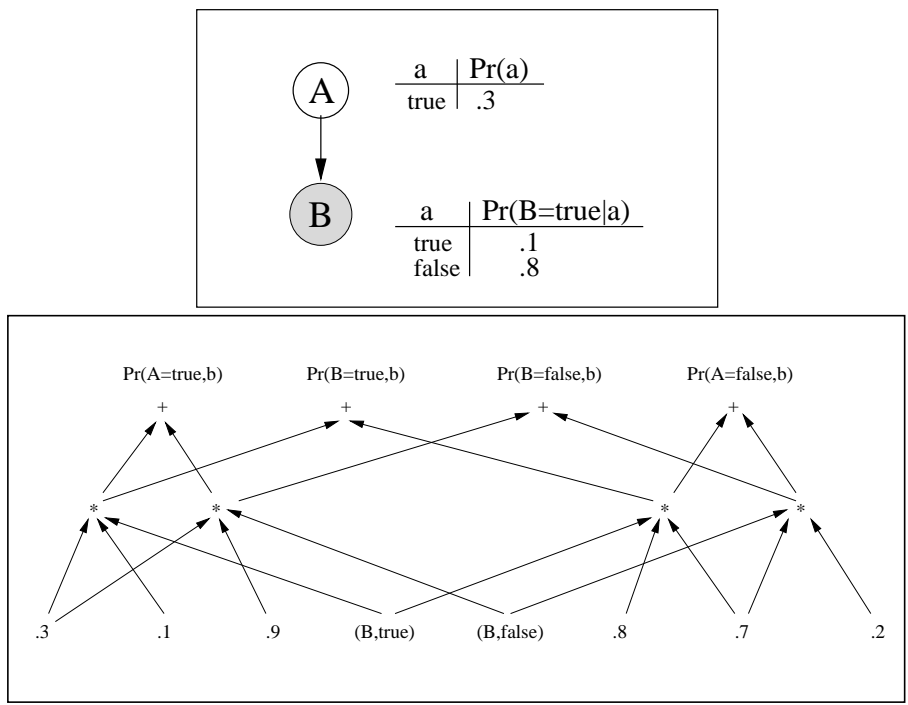

Figure 6: A belief network and its corresponding Q-DAG in which variable $B$ is declared to be both query and evidence.

and (2) applications where some variable may be expensive to observe, leading to an on-line decision on whether to observe it or not (using some value-of-information computation).

Both of these situations can be dealt with in a Q-DAG framework. First, as we mentioned earlier, Q-DAGs allow us to handle missing evidence through the use of the $\diamond$ notation which denotes an unknown value of a variable. Therefore, Q-DAGs can handle missing sensor readings. Second, a variable can be declared to be both query and evidence. This means that we can incorporate evidence about this variable when it is available, and also compute the probability distribution of the variable in case evidence is not available. Figure 6 depicts a Q-DAG in which variable $A$ is declared to be a query variable, while variable $B$ is declared to be both an evidence and a query variable (both variables have *true* and *false* as their values). In this case, we have two ESNs for variable $B$ and also two query nodes (see Figure 6). This Q-DAG can be used in two ways:

1. To compute the probability distributions of variables $A$ and $B$ when no evidence is available about $B$. Under this situation, the values of $n(B, true)$ and $n(B, false)$ are set to 1, and we have

$$
\begin{aligned}
Pr(A = true) &= .3 * .1 + .3 * .9 = .3 \\
Pr(A = false) &= .8 * .7 + .7 * .2 = .7 \\
Pr(B = true) &= .3 * .1 + .8 * .7 = .59
\end{aligned}
$$

$$Pr(B = false) = .3 * .9 + .7 * .2 = .41$$

2. To compute the probability of variable $A$ when evidence is available about $B$. For example, suppose that we observe $B$ to be *false*. The value of $n(B, true)$ will then be set to 0 and the value of $n(B, false)$ will be set to 1, and we have

$$Pr(A = true, B = false) = .3 * .9 = .27$$
$$Pr(A = false, B = false) = .7 * .2 = .14$$

The ability to declare a variable as both an evidence and a query variable seems to be essential in applications where (1) a decision may need to be made on whether to collect evidence about some variable $B$; and (2) making the decision requires knowing the probability distribution of variable $B$. For example, suppose that we are using the following formula (Pearl, 1988, Page 313) to compute the utility of observing variable $B$:

$$Utility\_Of\_Observing(B) = \sum_b Pr(B = b|\mathbf{e}) \, U(B = b),$$

where $U(B = b)$ is the utility for the decision maker of finding that variable $B$ has value $b$. Suppose that $U(B = true) = \$2.5$ and $U(B = false) = -\$3$. We can use the Q-DAG to compute the probability distribution of $B$ and use it to evaluate $Utility\_Of\_Observing(B)$:

$$Utility\_Of\_Observing(B) = (\$2.5 * .59) + (-\$3 * .41) = \$0.24,$$

which leads us to observe variable $B$. Observing $B$, we find that its value is *false*. We can then accommodate this evidence into the Q-DAG and continue with our analysis.

## 3. Generating Query DAGs

This section shows how Q-DAGs can be generated using traditional algorithms for exact belief-network inference. In particular, we will show how Q-DAGs can be generated using the clustering (join tree, Jensen, LS) algorithm (Jensen, Lauritzen, & Olesen, 1990; Shachter, Andersen, & Szolovits, 1994; Shenoy & Shafer, 1986), the polytree algorithm, and cutset conditioning (Pearl, 1988; Peot & Shachter, 1991). We will also outline properties that must be satisfied by other belief network algorithms in order to adapt them for generating Q-DAGs as we propose.

### 3.1 The Clustering Algorithm

We provide a sketch of the clustering algorithm in this section. Readers interested in more details are referred to (Shachter et al., 1994; Jensen et al., 1990; Shenoy & Shafer, 1986).

According to the clustering method, we start by:

1. constructing a join tree of the given belief network;[5]

---

5. A join tree is a tree of clusters that satisfies the following property: the intersection of any two clusters belongs to all clusters on the path connecting them.

2. assigning the matrix of each variable in the belief network to some cluster that contains the variable's family.

The join tree is a secondary structure on which the inference algorithm operates. We need the following notation to state this algorithm:

- $S_1, \ldots, S_n$ are the clusters, where each cluster corresponds to a set of variables in the original belief network.

- $\Psi_i$ is the *potential function* over cluster $S_i$, which is a mapping from instantiations of variables in $S_i$ into real numbers.

- $P_i$ is the *posterior probability distribution* over cluster $S_i$, which is a mapping from instantiations of variables in $S_i$ into real numbers.

- $M_{ij}$ is the message sent from cluster $S_i$ to cluster $S_j$, which is a mapping from instantiations of variables in $S_i \cap S_j$ into real numbers.

- $\mathbf{e}$ is the given evidence, that is, an instantiation of evidence variables $\mathbf{E}$.

We also assume the standard multiplication and marginalization operations on potentials.

Our goal now is to compute the potential $Pr(X, \mathbf{e})$ which maps each instantiation $x$ of variable $X$ in the belief network into the probability $Pr(x, \mathbf{e})$. Given this notation, we can state the algorithm as follows:

- Potential functions are initialized using

$$\Psi_i = \prod_X Pr_X \lambda_X,$$

  where

  - $X$ is a variable whose matrix is assigned to cluster $S_i$;
  - $Pr_X$ is the matrix for variable $X$: a mapping from instantiations of the family of $X$ into conditional probabilities; and
  - $\lambda_X$ is the likelihood vector for variable $X$: $\lambda_X(x)$ is 1 if $x$ is consistent with given evidence $\mathbf{e}$ and 0 otherwise.

- Posterior distributions are computed using

$$P_i = \Psi_i \prod_k M_{ki},$$

  where $S_k$ are the clusters adjacent to cluster $S_i$.

- Messages are computed using

$$M_{ij} = \sum_{S_i \backslash S_j} \Psi_i \prod_{k \neq j} M_{ki},$$

  where $S_k$ are the clusters adjacent to cluster $S_i$.

- The potential $Pr(X, \mathbf{e})$ is computed using

$$Pr(X, \mathbf{e}) = \sum_{S_i \setminus \{X\}} P_i,$$

where $S_i$ is a cluster to which $X$ belongs.

These equations are used as follows. To compute the probability of a variable, we must compute the posterior distribution of a cluster containing the variable. To compute the posterior distribution of a cluster, we collect messages from neighboring clusters. A message from cluster $S_i$ to $S_j$ is computed by collecting messages from all clusters adjacent to $S_i$ except for $S_j$.

This statement of the join tree algorithm is appropriate for situations where the evidence is not changing frequently since it involves computing initial potentials each time the evidence changes. This is not necessary in general and one can provide more optimized versions of the algorithm. This issue, however, is irrelevant in the context of generating Q-DAGs because updating probabilities in face of evidence changes will take place at the Q-DAG level, which includes its own optimization technique that we discuss later.

## 3.2 Generating Q-DAGs

To generate Q-DAGs using the clustering method, we have to go through two steps. First, we have to modify the initialization of potential functions so that the join tree is quantified using Q-DAG nodes instead of numeric probabilities. Second, we have to replace numeric addition and multiplication in the algorithm by analogous functions that operate on Q-DAG nodes. In particular:

1. Numeric multiplication $*$ is replaced by an operation $\otimes$ that takes Q-DAG nodes $n_1, \ldots, n_i$ as arguments, constructs and returns a new node $n$ with label $*$ and parents $n_1, \ldots, n_i$.

2. Numeric addition $+$ is replaced by an operation $\oplus$ that takes Q-DAG nodes $n_1, \ldots, n_i$ as arguments, constructs and returns a new node $n$ with label $+$ and parents $n_1, \ldots, n_i$.

Therefore, instead of numeric operations, we have Q-DAG-node constructors. And instead of returning a number as a computation result, we now return a Q-DAG node.

Before we state the Q-DAG clustering algorithm, realize that we now do not have evidence $\mathbf{e}$, but instead we have a set of evidence variables $\mathbf{E}$ for which we will collect evidence. Therefore, the Q-DAG algorithm will not compute an answer to a query $Pr(x, \mathbf{e})$, but instead will compute a Q-DAG node that evaluates to $Pr(x, \mathbf{e})$ under the instantiation $\mathbf{e}$ of variables $\mathbf{E}$.

In the following equations, potentials are mappings from variable instantiations to Q-DAG nodes (instead of numbers). For example, the matrix for variable $X$ will map each instantiation of $X$'s family into a Q-DAG node $n(p)$ instead of mapping it into the number $p$. The Q-DAG operations $\otimes$ and $\oplus$ are extended to operate on these new potentials in the same way that $*$ and $+$ are extended in the clustering algorithm.

The new set of equations is:

- Potential functions are initialized using

$$\Psi_i = \bigotimes_X n(Pr_X) \otimes \bigotimes_E n(\lambda_E),$$

where

- $X$ is a variable whose matrix is assigned to cluster $S_i$;
- $n(Pr_X)$ is the Q-DAG matrix for $X$: a mapping from instantiations of $X$'s family into Q-DAG nodes representing conditional probabilities;
- $E$ is an evidence variable whose matrix is assigned to cluster $S_i$; and
- $n(\lambda_E)$ is the Q-DAG likelihood vector of variable $E$: $n(\lambda_E)(e) = n(E, e)$, which means that node $n(\lambda_E)(e)$ evaluates to 1 if $e$ is consistent with given evidence and 0 otherwise.

- Posterior distributions are computed using

$$P_i = \Psi_i \bigotimes_k M_{ki},$$

where $S_k$ are the clusters adjacent to cluster $S_i$.

- Messages are computed using

$$M_{ij} = \bigoplus_{S_i \backslash S_j} \Psi_i \bigotimes_{k \neq j} M_{ki},$$

where $S_k$ are the clusters adjacent to cluster $S_i$.

- The Q-DAG nodes for answering queries of the form $Pr(x, \mathbf{e})$ are computed using

$$Qnode(X) = \bigoplus_{S_i \backslash \{X\}} P_i,$$

where $S_i$ is a cluster to which $X$ belongs.

Here $Qnode(X)$ is a potential that maps each instantiation $x$ of variable $X$ into the Q-DAG node $Qnode(X)(x)$ which evaluates to $Pr(x, \mathbf{e})$ for any given instantiation $\mathbf{e}$ of variables $\mathbf{E}$.

Hence, the only modifications we made to the clustering algorithm are (a) changing the initialization of potential functions and (b) replacing multiplication and addition with Q-DAG constructors of multiplication and addition nodes.

## 3.3 An Example

We now show how the proposed Q-DAG algorithm can be used to generate a Q-DAG for the belief network in Figure 4(a).

We have only one evidence variable in this example, $C$. And we are interested in generating a Q-DAG for answering queries about variable $B$, that is, queries of the form $Pr(b, \mathbf{e})$. Figure 7(a) shows the join tree for the belief network in Figure 4(a), where the tables contain the potential functions needed for the probabilistic clustering algorithm. Figure 7(b) shows the join tree again, but the tables contain the potential functions needed by the Q-DAG clustering algorithm. Note that the tables are filled with Q-DAGs instead of numbers.

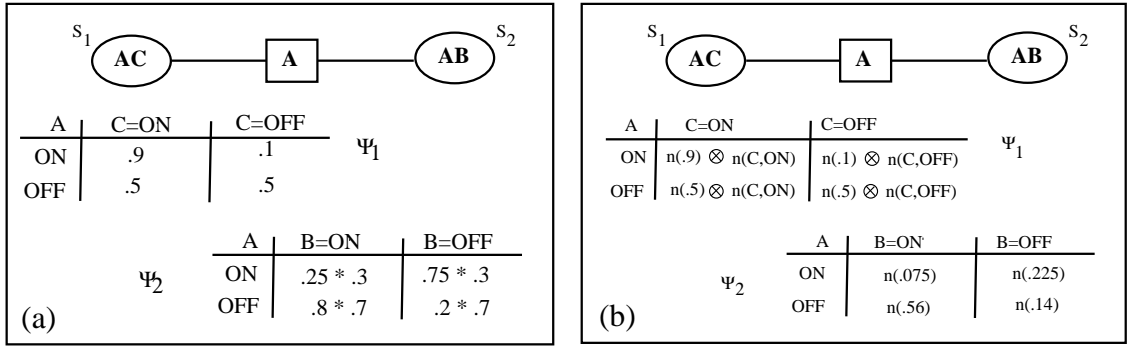

Figure 7: A join tree quantified with numbers (a), and with Q-DAG nodes (b).

We now apply the Q-DAG algorithm. To compute the Q-DAG nodes that will evaluate to $Pr(b, \mathbf{e})$, we must compute the posterior distribution $P_2$ over cluster $S_2$ since this is a cluster to which variable $B$ belongs. We can then sum the distribution over variable $A$ to obtain what we want. To compute the distribution $P_2$ we must first compute the message $M_{12}$ from cluster $S_1$ to cluster $S_2$.

The message $M_{12}$ is computed by summing the potential function $\Psi_1$ of cluster $S_1$ over all possible values of variable $C$, i.e., $M_{12} = \bigoplus_C \Psi_1$, which leads to:

$$M_{12}(A{=}ON) = [n(.9) \otimes n(C, ON)] \oplus [n(.1) \otimes n(C, OFF)],$$

$$M_{12}(A{=}OFF) = [n(.5) \otimes n(C, ON)] \oplus [n(.5) \otimes n(C, OFF)].$$

The posterior distribution over cluster $S_2$, $P_2$, is computed using $P_2 = \Psi_2 \otimes M_{12}$, which leads to

$$
\begin{aligned}
P_2(A{=}ON, B{=}ON) &= n(.075) \otimes [[n(.9) \otimes n(C, ON)] \oplus [n(.1) \otimes n(C, OFF)]] \\
P_2(A{=}ON, B{=}OFF) &= n(.225) \otimes [[n(.9) \otimes n(C, ON)] \oplus [n(.1) \otimes n(C, OFF)]] \\
P_2(A{=}OFF, B{=}ON) &= n(.56) \otimes [[n(.5) \otimes n(C, ON)] \oplus [n(.5) \otimes n(C, OFF)]] \\
P_2(A{=}OFF, B{=}OFF) &= n(.14) \otimes [[n(.5) \otimes n(C, ON)] \oplus [n(.5) \otimes n(C, OFF)]].
\end{aligned}
$$

The Q-DAG node $Qnode(b)$ for answering queries of the form $Pr(b, \mathbf{e})$ is computed by summing the posterior $P_2$ over variable $A$, $Qnode = \bigoplus_{S_2 \setminus \{B\}} P_2$, leading to

$$
\begin{aligned}
Qnode(B{=}ON) &= [n(.075) \otimes [[n(.9) \otimes n(C, ON)] \oplus [n(.1) \otimes n(C, OFF)]]] \oplus \\
&\quad [n(.56) \otimes [[n(.5) \otimes n(C, ON)] \oplus [n(.5) \otimes n(C, OFF)]]] \\
Qnode(B{=}OFF) &= [n(.225) \otimes [[n(.9) \otimes n(C, ON)] \oplus [n(.1) \otimes n(C, OFF)]]] \oplus \\
&\quad [n(.14) \otimes [[n(.5) \otimes n(C, ON)] \oplus [n(.5) \otimes n(C, OFF)]]],
\end{aligned}
$$

which is the Q-DAG depicted in Figure 4(b). Therefore, the result of applying the algorithm is two Q-DAG nodes, one will evaluate to $Pr(B{=}ON, \mathbf{e})$ and the other will evaluate to $Pr(B{=}OFF, \mathbf{e})$ under any instantiation $\mathbf{e}$ of evidence variables $\mathbf{E}$.

### 3.4 Computational Complexity of Q-DAG Generation

The computational complexity of the algorithm for generating Q-DAGs is determined by the computational complexity of the clustering algorithm. In particular, the proposed algorithm applies a $\oplus$-operation precisely when the clustering algorithm applies an addition-operation. Similarly, it applies a $\otimes$-operation precisely when the clustering algorithm applies a multiplication-operation. Therefore, if we assume that $\oplus$ and $\otimes$ take constant time, then both algorithms have the same time complexity.

Each application of $\oplus$ or $\otimes$ ends up adding a new node to the Q-DAG. And this is the only way a new node can be added to the Q-DAG. Moreover, the number of parents of each added node is equal to the number of arguments that the corresponding arithmetic operation is invoked on in the clustering algorithm. Therefore, the space complexity of a Q-DAG is the same as the time complexity of the clustering algorithm.

In particular, this means that the space complexity of Q-DAGs in terms of the number of evidence variables is the same as the time complexity of the clustering algorithm in those terms. Moreover, each evidence variable $E$ will add only $m$ evidence-specific nodes to the Q-DAG, where $m$ is the number of values that variable $E$ can take. This is important to stress because without this complexity guarantee it may be hard to distinguish between the proposed approach and a brute-force approach that builds a big table containing all possible instantiations of evidence variables together with their corresponding distributions of query variables.

### 3.5 Other Generation Algorithms

The polytree algorithm is a special case of the clustering algorithm as shown in (Shachter et al., 1994). Therefore, the polytree algorithm can also be modified as suggested above to compute Q-DAGs. This also means that cutset conditioning can be easily modified to compute Q-DAGs: for each instantiation $\mathbf{c}$ of the cutset $\mathbf{C}$, we compute a Q-DAG node for $Pr(x, \mathbf{c}, \mathbf{e})$ using the polytree algorithm and then take the $\oplus$-sum of the resulting nodes.

Most algorithms for exact inference in belief networks can be adapted to generate Q-DAGs. In general, an algorithm must satisfy a key condition to be adaptable for computing Q-DAGs as we suggested above. The condition is that the behavior of the algorithm should never depend on the specific evidence obtained, but should only depend on the variables about which evidence is collected. That is, whether variable $E$ is instantiated to value $v_1$ or value $v_2$ should not affect the complexity of the algorithm. Only whether variable $E$ is instantiated or not should matter.

Most belief networks algorithms that we are aware of satisfy this property. The reason for this seems to be the notion of probabilistic independence on which these algorithms are based. Specifically, what is read from the topology of a belief network is a relation $I(\mathbf{X}, \mathbf{Z}, \mathbf{Y})$, stating that variables $\mathbf{X}$ and $\mathbf{Y}$ are independent given variables $\mathbf{Z}$. That is,

$$Pr(\mathbf{x}, \mathbf{y} \mid \mathbf{z}) = Pr(\mathbf{x} \mid \mathbf{z}) Pr(\mathbf{y} \mid \mathbf{z})$$

for all instantiations $\mathbf{x}, \mathbf{y}, \mathbf{z}$ of these variables. It is possible, however, for this not to hold for all instantiations of $\mathbf{z}$ but only for specific ones. Most standard algorithms we are aware of do not take advantage of this instantiation–specific notion of independence.[6] Therefore, they cannot attach any computational significance to the specific value to which a variable is instantiated. This property of existing algorithms is what makes them easily adaptable to the generation of Q-DAGs.

### 3.6 Soundness of the Q-DAG Clustering Algorithm

The soundness of the proposed algorithm is stated below. The proof is given in Appendix A.

**Theorem 1** *Suppose that $Qnode(X)$ is a Q-DAG potential generated by the Q-DAG clustering algorithm for query variable $X$ and evidence variables $\mathbf{E}$. Let $\mathbf{e}'$ be an instantiation of some variables in $\mathbf{E}$, and let Q-DAG evidence $\mathcal{E}$ be defined as follows:*

$$\mathcal{E}(E) = \begin{cases} e, & \text{if evidence } \mathbf{e}' \text{ sets variable } E \text{ to value } e; \\ \diamond, & \text{otherwise.} \end{cases}$$

*then*

$$\mathcal{M}_{\mathcal{E}}(Qnode(X)(x)) = Pr(x, \mathbf{e}').$$

That is, the theorem guarantees that the Q-DAG nodes generated by the algorithm will always evaluate to their corresponding probabilities under any partial or full instantiation of evidence variables.

## 4. Reducing Query DAGs

This section is focused on reducing Q-DAGs after they have been generated. The main motivation behind this reduction is twofold: faster evaluation of Q-DAGs and less space to store them. Interestingly enough, we have observed that a few, simple reduction techniques tend in certain cases to subsume optimization techniques that have been influential in practical implementations of belief-network inference. Therefore, reducing Q-DAGs can be very important practically.

This section is structured as follows. First, we start by discussing four simple reduction operations in the form of rewrite rules. We then show examples in which these reductions subsume two key optimization techniques known as network-pruning and computation-caching.

### 4.1 Reductions

The goal of Q-DAG reduction is to reduce the size of a Q-DAG while maintaining the arithmetic expression it represents. In describing the equivalence of arithmetic expressions, we define the notion of Q-DAG equivalence:

**Definition 5** *Two Q-DAGs are <u>equivalent</u> iff they have the same set of evidence-specific nodes and they have the same output for all possible Q-DAG evidence.*

---

6. Some algorithms for two–level binary networks (BN20 networks), and some versions of the SPI algorithm do take advantage of these independences.

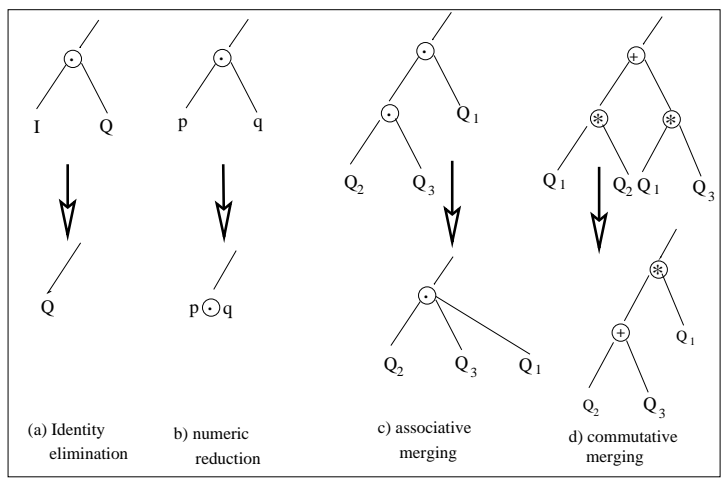

Figure 8: The four main methods for Q-DAG reduction.

Figure 8 shows four basic reduction operations that we have experimented with:

1. *Identity elimination:* eliminates a numeric node if it is an identity element of its child operation node.

2. *Numeric reduction:* replaces an operation node with a numeric node if all its parents are numeric nodes.

3. *Associative merging:* eliminates an operation node using operation associativity.

4. *Commutative merging:* eliminates an operation node using operation commutativity.

These rules can be applied successively and in different order until no more applications are possible.

We have proven that these operations are sound in (Darwiche & Provan, 1995). Based on an analysis of network structure and preliminary empirical results, we have observed that many factors govern the effectives of these operations. The degree to which reduction operations, numeric reduction in particular, can reduce the size of the Q-DAG depends on the topology of the given belief network and the set of evidence and query variables. For example, if all root nodes are evidence variables of the belief network, and if all leaf nodes are query variables, then numeric reduction will lead to little Q-DAG reduction.

We now focus on numeric reduction, showing how it sometimes subsumes two optimization techniques that have been influential in belief network algorithms. For both optimizations, we show examples where an unoptimized algorithm that employs numeric reduction yields the same Q-DAG as an optimized algorithm. The major implication is that optimizations can be done uniformly at the Q-DAG level, freeing the underlying belief network algorithms from such implementational complications.

The following examples assume that we are applying the polytree algorithm to singly-connected networks.

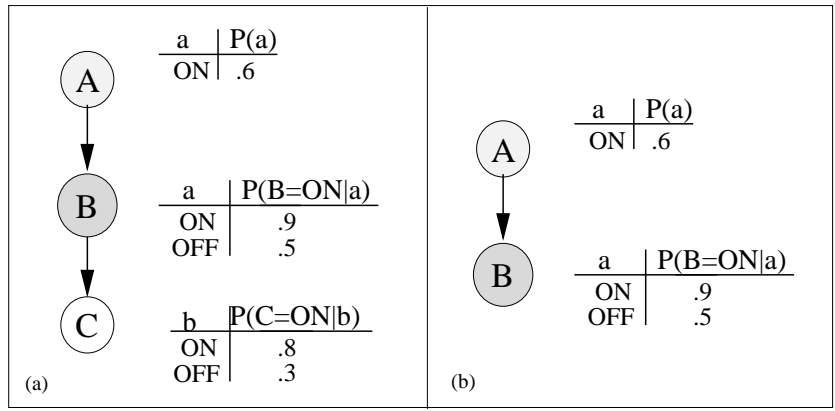

Figure 9: A simple belief network before pruning (a) and after pruning (b). The light-shaded node, $A$, is a query node, and the dark-shaded node, $B$, is an evidence node.

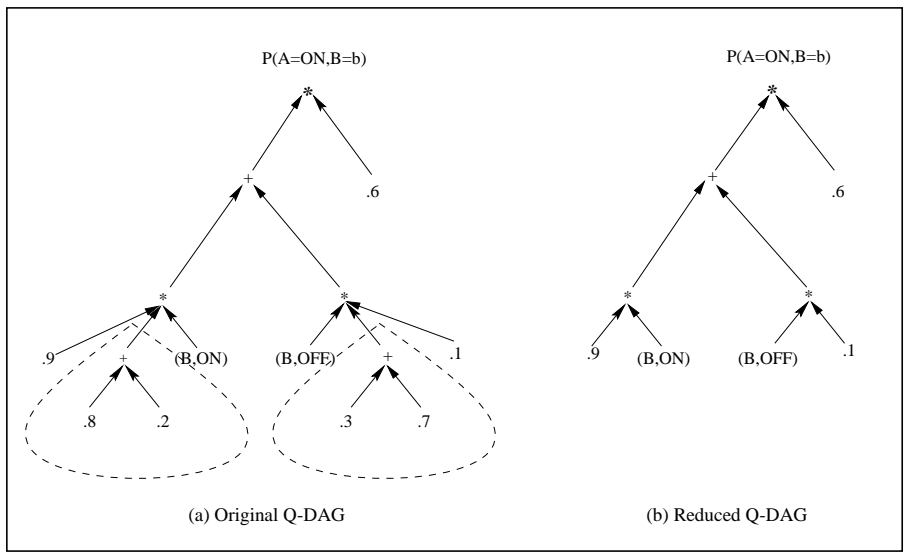

Figure 10: A Q-DAG (a) and its reduction (b).

## 4.2 Network Pruning

Pruning is the process of deleting irrelevant parts of a belief network before invoking inference. Consider the network in Figure 9(a) for an example, where $B$ is an evidence variable and $A$ is a query variable. One can prune node $C$ from the network, leading to the network in Figure 9(b). Any query of the form $Pr(a \mid b)$ has the same value with respect to either network. It should be clear that working with the smaller network is preferred. In general, pruning can lead to dramatic savings since it can reduce a multiply-connected network to a singly-connected one.

If we generate a Q-DAG for the network in Figure 9(a) using the polytree algorithm, we obtain the one in Figure 10(a). This Q-DAG corresponds to the following expression,

$$Pr(A{=}ON, \mathbf{e}) = Pr(A{=}ON) \sum_b \lambda_B(b) Pr(b \mid A{=}ON) \sum_c Pr(c \mid b).$$

If we generate a Q-DAG for the network in Figure 9(b), however, we obtain the one in Figure 10(b) which corresponds to the following expression,

$$Pr(A{=}ON, \mathbf{e}) = Pr(A{=}ON) \sum_b \lambda_B(b) Pr(b \mid A{=}ON).$$

As expected, this Q-DAG is smaller than the Q-DAG in Figure 10(a), and contains a subset of the nodes in Figure 10(a).

The key observation, however, is that the optimized Q-DAG in Figure 10(b) can be obtained from the unoptimized one in Figure 10(a) using Q-DAG reduction. In particular, the nodes enclosed in dotted lines can be collapsed using numeric reduction into a single node with value 1. Identity elimination can then remove the resulting node, leading to the optimized Q-DAG in Figure 10(b).

The more general observation, however, is that prunable nodes contribute identity elements when computing answers to queries. These contributions appear as Q-DAG nodes that evaluate to identity elements under all instantiations of evidence. Such nodes can be easily detected and collapsed into these identity elements using numeric reduction. Identity elimination can then remove them from the Q-DAG, leading to the same effect as network pruning.[7] Whether Q-DAG reduction can replace all possible pruning operations is an open question that is outside the scope of this paper.

### 4.3 Computation Caching

Caching computations is another influential technique for optimizing inference in belief networks. To consider an example, suppose that we are applying the polytree algorithm to compute $Pr(c, b)$ in the network of Figure 11. Given evidence, say $B{=}ON$, the algorithm will compute $Pr(c, B{=}ON)$ by passing the messages shown in Figure 12. If the evidence changes to $B{=}OFF$, however, an algorithm employing caching will not recompute the message $\pi_B(a)$ (which represents the causal support from $A$ to $B$ (Pearl, 1988)) since the value of this message does not depend on the evidence on $B$.[8] This kind of optimization is typically

---

7. Note, however, that Q-DAG reduction will not reduce the computational complexity of generating a Q-DAG, although network pruning may. For example, a multiply–connected network may become singly-connected after pruning, thereby, reducing the complexity of generating a Q-DAG. But using Q-DAG reduction, we still have to generate a Q-DAG by working with a multiply-connected network.

8. This can be seen by considering the following expression, which is evaluated incrementally by the polytree algorithm through its message passes:

$$Pr(c, \mathbf{e}) = \sum_b \underbrace{Pr(c \mid b) \lambda_B(b) \underbrace{\sum_a Pr(b \mid a) \underbrace{Pr(a)}_{\pi_B(a)}}_{\pi_C(b)}}.$$

It is clear that the subexpression corresponding to the message $\pi_B(a)$ from $A$ to $B$ is independent of the evidence on $B$.

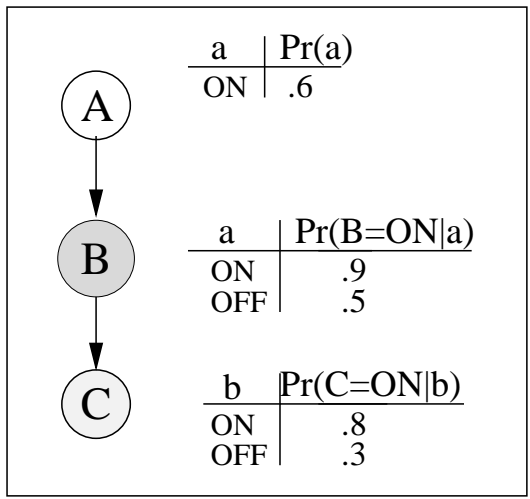

Figure 11: A simple belief network for demonstrating the relationship between Q-DAG reduction and computation caching. The light-shaded node, $C$, is a query node, and the dark-shaded node, $B$, is an evidence node.

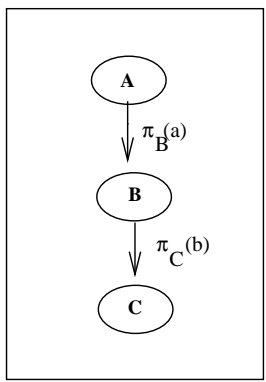

Figure 12: Message passing when $C$ is queried and $B$ is observed.

implemented by caching the values of messages and by keeping track of which messages are affected by what evidence.

Now, consider the Q-DAG corresponding to this problem which is shown in Figure 13(a). The nodes enclosed in dotted lines correspond to the message from $A$ to $B$.[9] These nodes do not have evidence-specific nodes in their ancestor set and, therefore, can never change values due to evidence changes. In fact, numeric reduction will replace each one of these nodes and its ancestors with a single node as shown in Figure 13(b).

In general, if numeric reduction is applied to a Q-DAG, one is guaranteed the following: (a) if a Q-DAG node represents a message that does not depend on evidence, that node will not be re-evaluated given evidence changes; and (b) numeric reduction will guarantee this under any Q-DAG evaluation method since it will replace the node and its ancestor set with a single root node.[10]

---

9. More precisely, they correspond to the expression $\sum_a Pr(b \mid a) Pr(a)$.

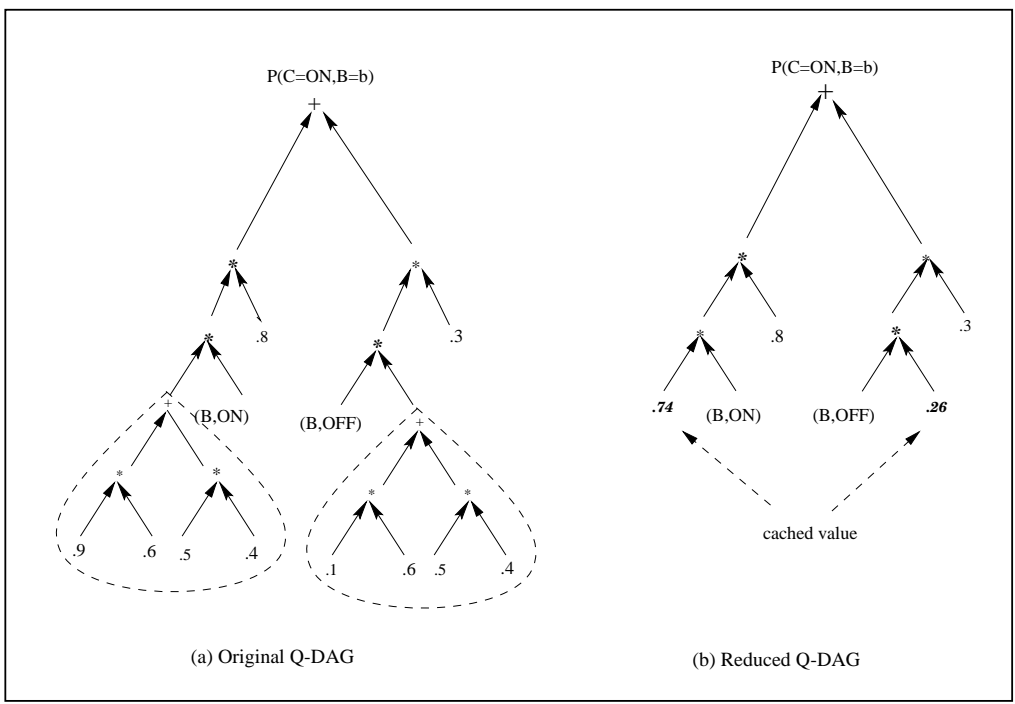

Figure 13: A Q-DAG (a) and its reduction (b).

## 4.4 Optimization in Belief-Network Inference

Network pruning and computation caching have proven to be very influential in practical implementations of belief-network inference. In fact, our own experience has shown that these optimizations typically make the difference between a usable and a non-usable belief-network system.

One problem with these optimizations, however, is their algorithm-specific implementations although they are based on general principles (e.g., taking advantage of network topology). Another problem is that they can make elegant algorithms complicated and hard to understand. Moreover, these optimizations are often hard to define succinctly, and hence are not well documented within the community.

In contrast, belief–network inference can be optimized by generating Q-DAGs using un-optimized inference algorithms, and then optimizing the generated Q-DAG through reduction techniques. We have shown some examples of this earlier with respect to pruning and caching optimizations. However, whether this alternate approach to optimization is always feasible is yet to be known. A positive answer will clearly provide *an algorithm–independent*

---

10. Note that Q-DAGs lead to a very refined caching mechanism if the Q-DAG evaluator (1) caches the value of each Q-DAG node and (2) updates these cached values only when there is need to (that is, when the value of a parent node changes). Such a refined mechanism allows caching the values of messages that *depend* on evidence as well.

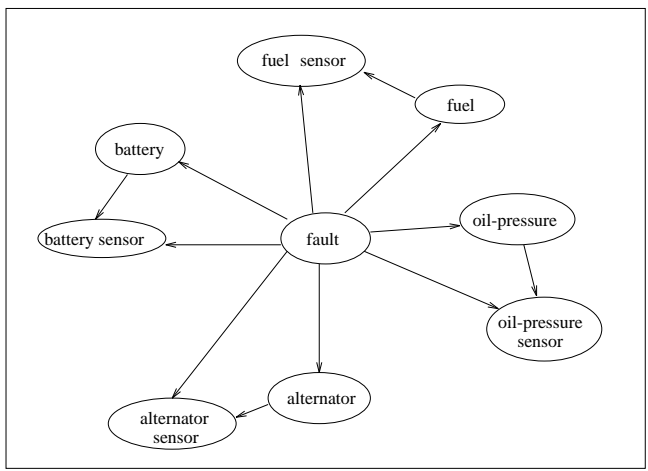

Figure 14: A simple belief network for car diagnosis.

*approach to optimizing belief–network inference*, which is practically important for at least two reasons. First, Q-DAG reduction techniques seem to be much simpler to understand and implement since they deal with graphically represented arithmetic expressions, without having to invoke probability or belief network theory. Second, reduction operations are applicable to Q-DAGs generated by any belief–network algorithm. Therefore, an optimization approach based on Q-DAG reduction would be more systematic and accessible to a bigger class of developers.

## 5. A Diagnosis Example

This section contains a comprehensive example illustrating the application of the Q-DAG framework to diagnostic reasoning.

Consider the car troubleshooting example depicted in Figure 14. For this simple case we want to determine the probability distribution for the fault node, given evidence on four sensors: the battery-, alternator-, fuel- and oil-sensors. Each sensor provides information about its corresponding system. The fault node defines five possible faults: normal, clogged-fuel-injector, dead-battery, short-circuit, and broken-fuel-pump.

If we denote the fault variable by $F$, and sensor variables by $\mathbf{E}$, then we want to build a system that can compute the probability $Pr(f, \mathbf{e})$, for each fault $f$ and any evidence $\mathbf{e}$. These probabilities represent an unnormalized probability distribution over the fault variable given sensor readings. In a Q-DAG framework, realizing this diagnostic system involves three steps: Q-DAG generation, reduction, and evaluation. The first two steps are accomplished off-line, while the final step is performed on-line. We now discuss each one of the steps in more detail.

### 5.1 Q-DAG Generation

The first step is to generate the Q-DAG. This is accomplished by applying the Q-DAG clustering algorithm with the fault as a query variable and the sensors as evidence vari-ables. The resulting Q-DAG has five query nodes, $Qnode(F = normal, \mathbf{e})$, $Qnode(F = clogged\_fuel\_injector, \mathbf{e})$, $Qnode(F = dead\_battery, \mathbf{e})$, $Qnode(F = short\_circuit, \mathbf{e})$, and $Qnode(F = broken\_fuel\_pump, \mathbf{e})$. Each node evaluates to the probability of the corresponding fault under any instantiation of evidence. The probabilities constitute a differential diagnosis that tells us which fault is most probable given certain sensor values.

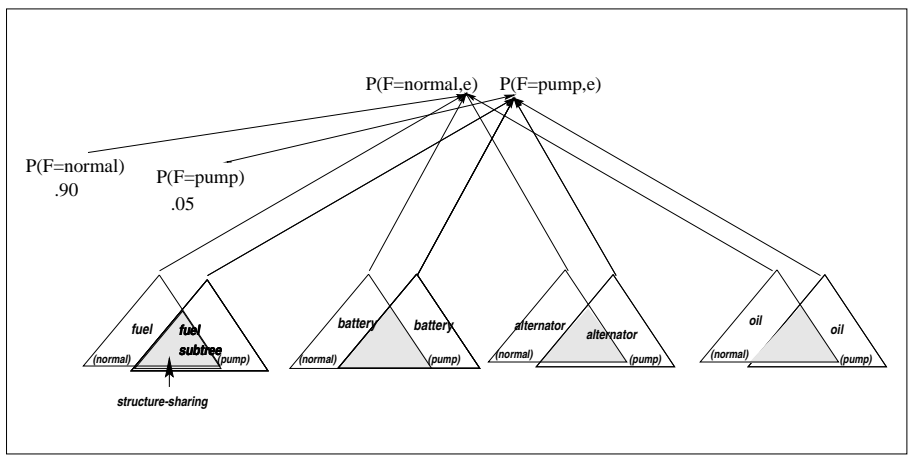

Figure 15: A partial Q-DAG for the car example, displaying two of the five query nodes, *broken_fuel_pump* and *normal*. The shaded regions are portions of the Q-DAG that are shared by multiple query nodes; the values of these nodes are relevant to the value of more than one query node.

Figure 15 shows a stylized description of the Q-DAG restricted to two of the five query nodes, corresponding to $Pr(F = broken\_fuel\_pump, \mathbf{e})$ and $Pr(F = normal, \mathbf{e})$. The Q-DAG structure is symmetric for each fault value and sensor.

Given that the Q-DAG is symmetric for these possible faults, for clarity of exposition we look at just the subset needed to evaluate node $Pr(F = broken\_fuel\_pump, \mathbf{e})$. Figure 16 shows a stylized version of the Q-DAG produced for this node. Following are some observations about this Q-DAG. First, there is an evidence-specific node for every instantiation of sensor variables, corresponding to all forms of sensor measurements possible. Second, all other roots of the Q-DAG are probabilities. Third, one of the five parents of the query node $Pr(F = broken\_fuel\_pump, \mathbf{e})$ is for the prior on $F = broken\_fuel\_pump$, and the other four are for the contributions of the four sensors. For example, Figure 16 highlights (in dots) that part of the Q-DAG for computing the contribution of the battery sensor.

## 5.2 Q-DAG Reduction

After generating a Q-DAG, one proceeds by reducing it using graph rewrite rules. Figure 16 shows an example of such reduction with a Q-DAG that is restricted to one query node for simplicity. To give an idea of the kind of reduction that has been applied, consider the partial Q-DAG enclosed by dots in this figure. Figure 17 compares this reduced Q-DAG with the unreduced one from which it was generated. Given our goal of generating Q-DAGs that (a) can be evaluated as efficiently as possible and (b) require minimal space to store, it is important to see, even in a simple example, how Q-DAG reduction can make a big difference in their size.

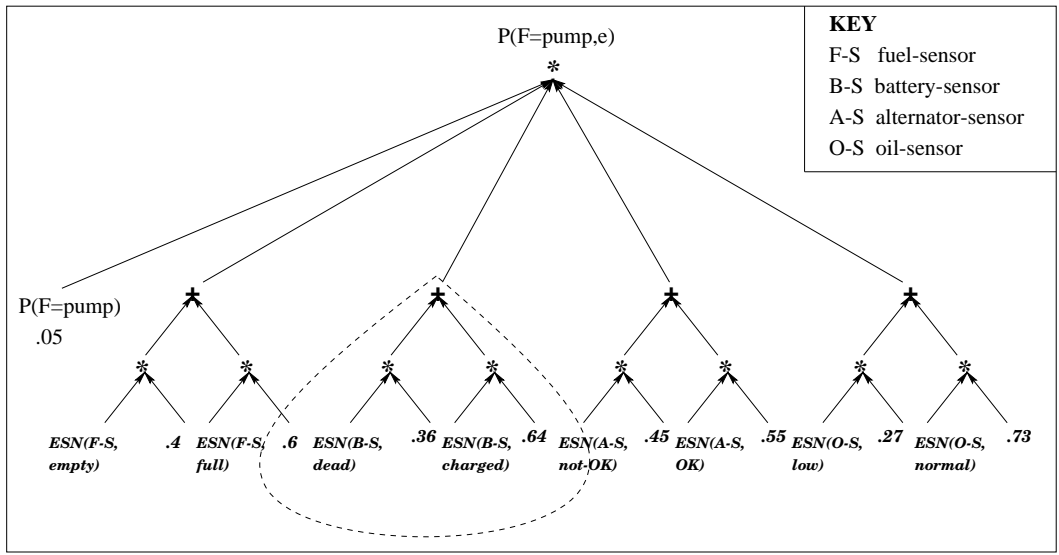

Figure 16: A partial Q-DAG for the car example.

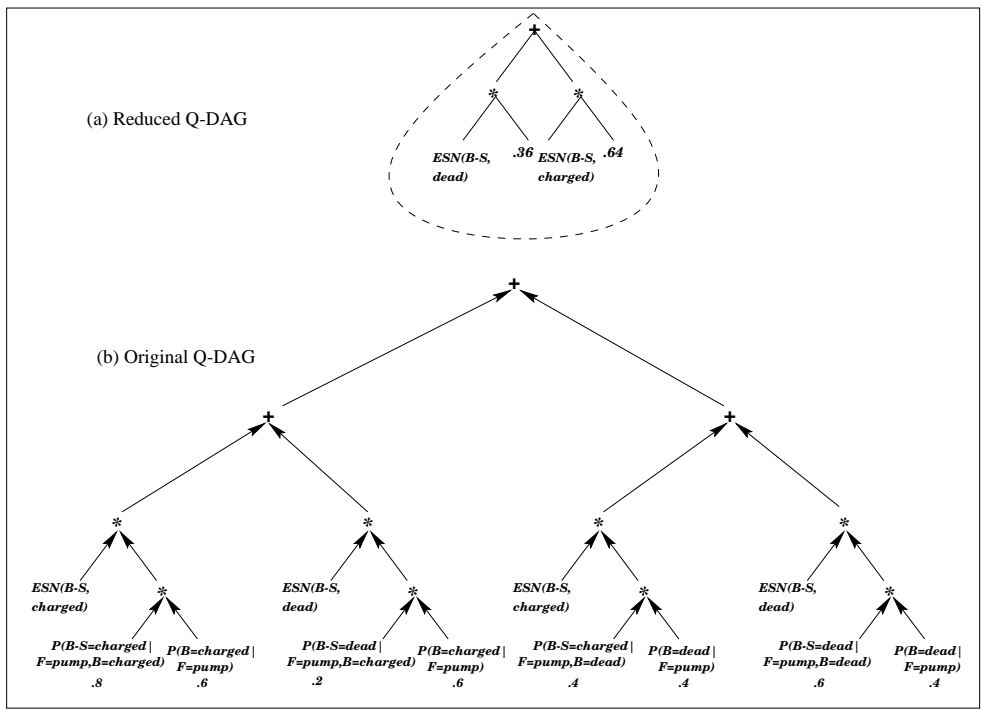

Figure 17: Reduced and unreduced Q-DAGs for the car diagnosis example.

### 5.3 Q-DAG Evaluation

Now that we have a reduced Q-DAG, we can use it to compute answers to diagnostic queries. This section presents examples of this evaluation with respect to the generated Q-DAG.

Suppose that we obtain the readings dead, normal, ok and full for the battery, oil, alternator and fuel sensors, respectively. And let us compute the probability distribution over the fault variable. This obtained evidence is formalized as follows:

- $\mathcal{E}(battery\_sensor) = dead$,

- $\mathcal{E}(oil\_sensor) = normal$,

- $\mathcal{E}(alternator\_sensor) = ok$,

- $\mathcal{E}(fuel\_sensor) = full$.

Evidence-specific nodes can now be evaluated according to Definition 3. For example, we have

$$\mathcal{M}_{\mathcal{E}}[n(battery\_sensor, charged)] = 0,$$

and

$$\mathcal{M}_{\mathcal{E}}[n(battery\_sensor, dead)] = 1.$$

The evaluation of evidence-specific nodes is shown pictorially in Figure 18(a). Definition 3 can then be used to evaluate the remaining nodes: once the values of a node's parents are known, the value of that node can be determined. Figure 18(b) depicts the results of evaluating other nodes. The result of interest here is the probability 0.00434 assigned to the query node $Pr(fault = broken\_fuel\_pump, \mathbf{e})$.

Suppose now that evidence has changed so that the value of fuel sensor is empty instead of full. To update the probability assigned to node $Pr(fault = broken\_fuel\_pump, \mathbf{e})$, a brute force method will re-evaluate the whole Q-DAG. However, if a forward propagation scheme is used to implement the node evaluator, then only four nodes need to be re-evaluated in Figure 18(b) (those enclosed in circles) instead of thirteen (the total number of nodes). We stress this point because this refined updating scheme, which is easy to implement in this framework, is much harder to achieve when one attempts to embed it in standard belief-network algorithms based on message passing.

## 6. Concluding Remarks

We have introduced a new paradigm for implementing belief-network inference that is oriented towards real-world, on-line applications. The proposed framework utilizes knowledge of query and evidence variables in an application to compile a belief network into an arithmetic expression called a Query DAG (Q-DAG). Each node of a Q-DAG represents a numeric operation, a number, or a symbol that depends on available evidence. Each leaf node of a Q-DAG represents the answer to a network query, that is, the probability of some event of interest. Inference on Q-DAGs is linear in their size and amounts to a standard evaluation of the arithmetic expressions they represent.

A most important point to stress about the work reported here is that it is *not* proposing a new algorithm for belief-network inference. What we are proposing is a paradigm for implementing belief-network inference that is orthogonal to standard inference algorithms and is engineered to meet the demands of real-world, on-line applications. This class of applications is typically demanding for the following reasons:

1. It typically requires very short response time, i.e., milliseconds.

2. It requires software to be written in specialized languages, such as ADA, C++, and assembly before it can pass certification procedures.

3. It imposes severe restrictions on the available software and hardware resources in order to keep the cost of a "unit" (such as an electromechanical device) as low as possible.

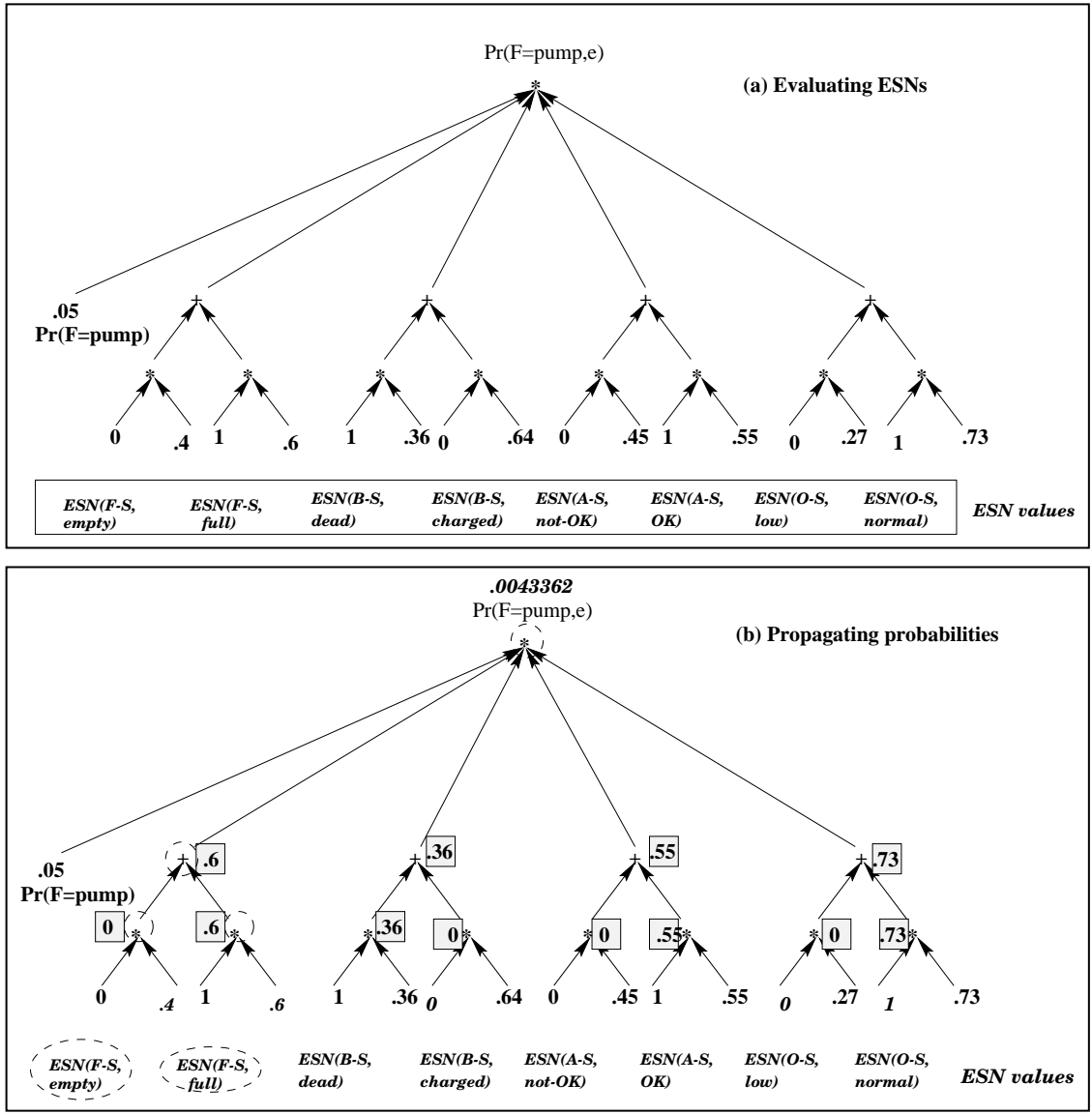

Figure 18: Evaluating the Q-DAG for the car diagnosis example given evidence for sensors. The bar in (a) indicates the instantiation of the ESNs. The shaded numbers in (b) indicate probability values that are computed by the node evaluator. The circled operations on the left-hand-side of (b) are the only ones that need to be updated if evidence for the fuel-system sensor is altered, as denoted by the circled ESNs.

To address these real-world constraints, we are proposing that one compile a belief network into a Q-DAG as shown in Figure 3 on and use a Q-DAG evaluator for on-line reasoning. This brings down the required memory to that needed for storing a Q-DAG and its evaluator. It also brings down the required software to that needed for implementing a Q-DAG evaluator, which is very simple as we have seen earlier.

Our proposed approach still requires a belief-network algorithm to generate a Q-DAG, but it makes the efficiency of such an algorithm less of a critical factor.[11] For example, we show that some standard optimizations in belief-network inference, such as pruning and caching, become less critical in a Q-DAG framework since these optimizations tend to be subsumed by simple Q-DAG reduction techniques, such as numeric reduction.

The work reported in this paper can be extended in at least two ways. First, further Q-DAG reduction techniques could be explored, some oriented towards reducing the evaluation time of Q-DAGs, others towards minimizing the memory needed to store them. Second, we have shown that some optimization techniques that dramatically improve belief-network algorithms may become irrelevant to the size of Q-DAGs if Q-DAG reduction is employed. Further investigation is needed to prove formal results and guarantees on the effectiveness of Q-DAG reduction.

We close this section by noting that the framework we proposed is also applicable to order-of-magnitude (OMP) belief networks, where multiplication and addition get replaced by addition and minimization, respectively (Goldszmidt, 1992; Darwiche & Goldszmidt, 1994). The OMP Q-DAG evaluator, however, is much more efficient than its probabilistic counterpart since one may evaluate a minimization node without having to evaluate all its parents in many cases. This can make considerable difference in the performance of a Q-DAG evaluator.

## Acknowledgements

Most of the work in this paper was carried out while the first author was at Rockwell Science Center. Special thanks to Jack Breese, Bruce D'Ambrosio and to the anonymous reviewers for their useful comments on earlier drafts of this paper.

---

11. We have shown how clustering and conditioning algorithms can be used for Q-DAG generation, but other algorithms such as SPI (Li & D'Ambrosio, 1994; Shachter et al., 1990) can be used as well.

## Appendix A. Proof of Theorem 1

Without loss of generality, we assume in this proof that all variables are declared as evidence variables. To prove this soundness theorem, all we need to show is that each Q-DAG potential will evaluate to its corresponding probabilistic potential under all possible evidence. Formally, for any cluster $S$ and variables $X$, the matrices of which are assigned to $S$, we need to show that

$$\mathcal{M}_{\mathcal{E}}(\bigotimes_X n(Pr_X) \otimes n(\lambda_X)) = \prod_X Pr_X \lambda_X \tag{1}$$

for a given evidence $\mathcal{E}$. Once we establish this, we are guaranteed that $Qnode(X)(x)$ will evaluate to the probability $Pr(x, \mathbf{e})$ because the application of $\otimes$ and $\oplus$ in the Q-DAG algorithm is isomorphic to the application of $*$ and $+$ in the probabilistic algorithm, respectively.

To prove Equation 1, we will extend the Q-DAG node evaluator $\mathcal{M}_{\mathcal{E}}$ to mappings in the standard way. That is, if $f$ is a mapping from instantiations to Q-DAG nodes, then $\mathcal{M}_{\mathcal{E}}(f)$ is defined as follows:

$$\mathcal{M}_{\mathcal{E}}(f)(x) =_{def} \mathcal{M}_{\mathcal{E}}(f(x)).$$

That is, we simply apply the Q-DAG node evaluator to the range of mapping $f$.

Note that $\mathcal{M}_{\mathcal{E}}(f \otimes g)$ will then be equal to $\mathcal{M}_{\mathcal{E}}(f)\mathcal{M}_{\mathcal{E}}(g)$. Therefore,

$$\mathcal{M}_{\mathcal{E}}(\bigotimes_X n(Pr_X) \otimes n(\lambda_X))$$
$$= \prod_X \mathcal{M}_{\mathcal{E}}(n(Pr_X))\mathcal{M}_{\mathcal{E}}(n(\lambda_X))$$
$$= \prod_X Pr_X \mathcal{M}_{\mathcal{E}}(n(\lambda_X)) \text{ by definition of } n(Pr_X).$$

Note also that by definition of $n(\lambda_X)$, we have that $n(\lambda_X)(x)$ equals $n(X, x)$. Therefore,

$$
\begin{aligned}
\mathcal{M}_{\mathcal{E}}(n(\lambda_X))(x) &= \mathcal{M}_{\mathcal{E}}(n(\lambda_X)(x)) \\
&= \mathcal{M}_{\mathcal{E}}(n(X, x)) \\
&= \begin{cases} 1, & \text{if } \mathcal{E}(X) = x \text{ or } \mathcal{E}(X) = \diamond \\ 0, & \text{otherwise} \end{cases} \\
&= \lambda_X(x).
\end{aligned}
$$

Therefore,

$$\mathcal{M}_{\mathcal{E}}(\bigotimes_X n(Pr_X) \otimes n(\lambda_X)) = \prod_X Pr_X \lambda_X. \quad \blacksquare$$